\documentclass{article}

\usepackage{arxiv}
\usepackage{cite}
\usepackage{amsmath,amssymb,amsfonts}
\usepackage{algorithmic}
\usepackage{graphicx}
\usepackage{textcomp}
\usepackage{xcolor}
\usepackage{hyperref}
\usepackage{multicol}
\usepackage{algorithm}
\usepackage{algorithmic}
\usepackage{hyperref}
\hypersetup{
    colorlinks=false,
    linkcolor=blue,
    filecolor=magenta,      
    urlcolor=cyan,
}
\def\BibTeX{{\rm B\kern-.05em{\sc i\kern-.025em b}\kern-.08em
    T\kern-.1667em\lower.7ex\hbox{E}\kern-.125emX}}
    
\pdfoutput=1

\begin{document}

\title{TinyFedTL: Federated Transfer Learning on Tiny Devices\\
}

\author{ {Kavya Kopparapu *} \\
	Department of Computer Science\\
	Harvard University\\
	Cambridge, MA, USA \\
	\texttt{kavyakopparapu@college.harvard.edu} \\
	\And
	{Eric Lin *} \\
	Department of Computer Science\\
	Harvard University\\
	Cambridge, MA, USA \\
	\texttt{eric\_lin@college.harvard.edu} \\
}

\thanks{Authors contributed equally to this research}

\maketitle

\begin{abstract}
TinyML has rose to popularity in an era where data is everywhere. However, the data that is in most demand is subject to strict privacy and security guarantees. In addition, the deployment of TinyML hardware in the real world has significant memory and communication constraints that traditional ML fails to address. In light of these challenges, we present TinyFedTL, the first implementation of federated transfer learning on a resource-constrained microcontroller. 
\end{abstract}


\section{Introduction}
Our presentation is available at \href{https://harvard.zoom.us/rec/share/xolCUYS-w-MmXAfRVWHl71b8HcpewsKCL7hqSAVlEy4G9kv-iC4Xk06acVW3oFQ5.XGkN1q3kZp51U_Uv?startTime=1608180080000}{this link}.

In recent years, the emphasis on data privacy has grown in the wake of several privacy scandals and information leaks. More than ever, individuals are concerned with who has access to their personal data and where it is being shared. Most current successful machine learning methods, for both the Tiny domain and not, benefit from large and diverse datasets, so this rise in concern regarding digital privacy appears to come at the cost of progress.

In addition, in many applications there is a significant need for distributed learning agents that operate in an environment with significant communication costs and minimal on-device storage. These distributed computing setups have even taken hold in consumer-facing applications such as the Amazon Go store. Tiny Machine Learning is a rapidly growing field at the intersection of embedded systems and machine learning, allowing significant insights, data collection, and algorithmic development that was not previously possible. 

Put together, the opportunity is twofold: data from several sources can no longer be consolidated for a singular learner to access due to privacy concerns and low-cost learning devices on the edge need a new method to aggregate shared insights that doesn’t require large on-device memory and constant communication to a central server. Thus, we see the field of TinyML is ripe for applications of privacy-preserving machine learning. However, frameworks like Tensorflow Lite do not currently support model training on-device, but rather enable the deployment of static models for inference \cite{david2020tensorflow}. 

In this work, we contribute:
\begin{itemize}
    \item The first implementation of federated learning on a microcontroller.
    \item A method of deploying transfer learning on a resource-constrained microconroller without growing storage costs as the number of training examples increases. 
    \item The method of federated transfer learning for the CIFAR-10 benchmark doesn't necessitate a dense layer with pre-trained weights, meaning devices can be re-trained to different types of classification problems. 
    \item Implemmentation of federated transfer learning on an Arduino microcontroller. 
    \item Identification of challenges and limitations encountered in the process of training on-device with a federated learning framework.
\end{itemize}

\section{Related Work}
\subsection{TinyML}
Previous work in TinyML is focused on optimizations to compress models and reduce the inference latency but does not aim to allow continuous learning, especially in a privacy-preserving manner. Prior research in deploying learning on-device in the TinyML domain has included simple NN classifiers like k-Nearest Neighbors to transfer learn on-device \cite{disabato2020incremental}. However, such a classifier scales poorly with more training examples and does not follow federated learning privacy guarantees, as information about the input to a network can be generated from the embedding generated without the final dense layer. Therefore, such an implementation is essentially infeasible on a hardware platform such as the Arduino Nano. 

\subsection{Federated Learning}
The domain of Federated Learning has been a hot research topic in recent years. The topic first emerged in late 2015, with the seminal paper by McMahan et al. of Google AI detailing their novel approach and proposing a concrete “Federated averaging” algorithm \cite{mcmahan2017communication}. The introduction of this algorithm presented a solution to several bottlenecks concerning mobile devices and privacy constraints. By distributing the data across edge devices, user privacy can be respected while learning models can be conducted on the aggregate updates on the collection of devices \cite{kairouz2019advances}. There has been continued work in addressing problems in Federated Learning, including non-iid data distribution, attacks, and communication cost \cite{balcan2012distributed, shoham2019overcoming, 2020fedcd}.

\subsection{FL for TinyML}
Due to the fact that both fields are relatively new, there has been little work conducted in the intersection of tinyML and Federated Learning. There has been some work in the implementation of Federated Learning in mobile edge network and in selectively updating parts of large networks to make transfer learning more feasible but none have integrated into a significantly constrained device, but instead onto a Raspberry Pi \cite{lim2020federated, cai2020tinytl}.

\section{Methods and Experiments}
\subsection{Transfer Learning Task}
In order to turn a CNN into a feature extractor, we removed the final fully connected (dense) layer, so the output of the CNN was a feature vector $n$ units long. We took the weights for this feature vector off-the-shelf, meaning we did not train the model on any specific task but rather kept the weights from a large dataset that were used to generate meaningful features (in this case, the dataset was either ImageNet or Visual Wake Words) \cite{deng2009imagenet, chowdhery2019visual}. We then used the features extracted by the CNN extractor to be the input into a fully connected layer, whose outputs were then put through a softmax to get class probabilities. Training the weights of this FC on-device later but not the feature extractor (which is set aside as constant) is the task.

A typical benchmark for Federated Learning is the CIFAR-10 dataset, which consists of $32$x$32$ images in 10 different classes. These images were upscaled to $96$x$96$ and turned into two binary classification problems: dog versus no-dog and cat versus no-cat. The dog classification problem was treated as the original problem (for optionally pre-training the the dense layer of the off-the-shelf model trained on ImageNet or Visual Wake Words) and the cat classification problem was treated as the transfer problem.
 
\subsection{Hardware}
The microcontroller of choice for these experiments was the Arduino Nano 33 BLE Sense. The BLE Sense is the hardware of choice for many TinyML applications due to its variety of sensors (including temperature, pressure, humidity, light, color, and more) and interface with the Arduino IDE. 

The Arduino federated learning implementation also used a 5MP Arducam to collect real-world images as local training data. A macbook simulated a global server by communicating with the Arduino through a serial port. Finally, unlike other approaches for on-device learning, no SD card or external storage was used in simulations. This meant that all training and data storage had to take place in the 1MB of flash memory and 256KB of SRAM, further constraining the memory capacity.

\subsection{Model Transfer Learning On-Device}
Due to the incompatibility of the current version of Tensorflow Lite Micro with on-device training and update techniques, we implemented our own fully connected (FC) layer inference and backpropogation update in C++. 

\subsection{Arduino Federated Learning Implementation}
The Federated Averaging algorithm, the gold standard for enabling federated learning across distributed devices, is shown in Algorithm \ref{algo}. This algorithm is the basis for our TinyFedTL system. 
\begin{algorithm}[h]
\caption{FedAvg Algorithm \cite{mcmahan2017communication}}
\label{alg}
\begin{algorithmic}
    \STATE {\bfseries Input:} Devices $i = 1, ..., N$
    \FOR{epoch $t = 1, 2, \dots,T$}
            \FOR {$i \in \textit{N}$}
                \STATE Device $i$ trains all models $m$ on its local data for $E$ local episodes
            \ENDFOR
            \FOR{$m = 1,2, \dots, M$}
                \STATE $\textit{w\_avg} = \text{AverageWeights}(i \text{ s.t. } c_m^{(i)} \neq 0)$\\
                \STATE Learner updates model $m$ with  \textit{w\_avg}\\
            \ENDFOR
            \STATE  Evaluate models with global validation data
    \ENDFOR
\end{algorithmic}
\label{algo}
\end{algorithm}

Federated Learning in the real-world likely only sees a given example or data point once. This is notably different than normal federated (or not) training schemes, in which training examples are reused over epochs of training. Furthermore, theoretical federated learning schemes push and send weight updates on a carefully scheduled and consistent basis. This is simply infeasible in many applications of tinyML edge devices where unstable network connection may prevent comunication for long stretches of time. Moreover, data input is not uniform -- changing environments may result in some days where a certain edge device sees many pieces of new data, while other devices see none.

Our setup mirrors real-world scenarios as described above. Data available to the model is different for every epoch of training and collected through the Arducam. Furthermore, the arduino continuously sees and trains on new data until it is contacted by the global server to update its weights. When the server does initiate contact, model weight and bias updates have to be sent and read byte by byte via the serial port between the arduino and the server. The batch size on our real-world implementation is 1 (instantaneous inference and model update). 

\section{Results}
\subsection{Models}
We implemented two versions of MobileNetV2 for our task: the Tensorflow built-in MobileNet compressed with a factor of $alpha = 0.35$ (hereby referenced as tf-mobilenet) pretrained on ImageNet and the MobileNet from the TinyML Perf Benchmark (hereby referenced as perf-mobilenet) \cite{deng2009imagenet, banbury2020benchmarking}. The compressed tf-mobilenet has more parameters than perf-mobilenet has x parameters. Both models were frozen and the input and outputs were quantized using post-training 8-bit quantization to interface correctly with the microcontroller.  

\begin{table}[htbp]
\caption{Deployed Model Descriptions}
\begin{center}
\begin{tabular}{|c|c|c|c|c|}
\hline
\textbf{Model}&\multicolumn{4}{|c|}{\textbf{Model Parameters}} \\
\cline{2-5} 
\textbf{Name} & \textbf{\textit{Total}}& \textbf{\textit{Trainable}}& \textbf{\textit{Trainable}}& \textbf{\textit{Trained on}} \\
\hline
tf-mobilenet& 412,770& 2,562& 410,208& ImageNet\\
perf-mobilenet& 221,794& 514& 221,280& Visual Wake Words\\
\hline
\end{tabular}
\label{tab1}
\end{center}
\end{table}

We ran into issues with transfer learning on the perf-mobilenet model. Specifically, the embeddings generated from the network were incredibly sparse, where only 16 of the 256 outputs were ever non-zero. This may be as a result of encouraging sparsity during training via methods such as L2 weight regularization so the embeddings generated are highly specific to the task it was initially trained on (person detection through the visual wake words dataset). If given more time, we would have re-trained the perf-mobilenet to get more meaningful embeddings.

In our simulations, tf-mobilenet performed much better and successfully learned the task of cat identification. However, it was too large to fit on-device for our microcontroller. Since we were using an Arduino Nano 33 BLE Sense with 1MB of flash memory and 256KB of SRAM, we believe that other microcontrollers could have fit the tf-mobilenet model. Therefore, we used the perf-mobilenet for memory and time benchmarking on our arduino and the tf-mobilenet model in our simulations to understand performance. 
\subsection{Transfer Learning from Scratch versus Pretrained}
\begin{figure}[htbp]
\centerline{\includegraphics[width=8.5cm]{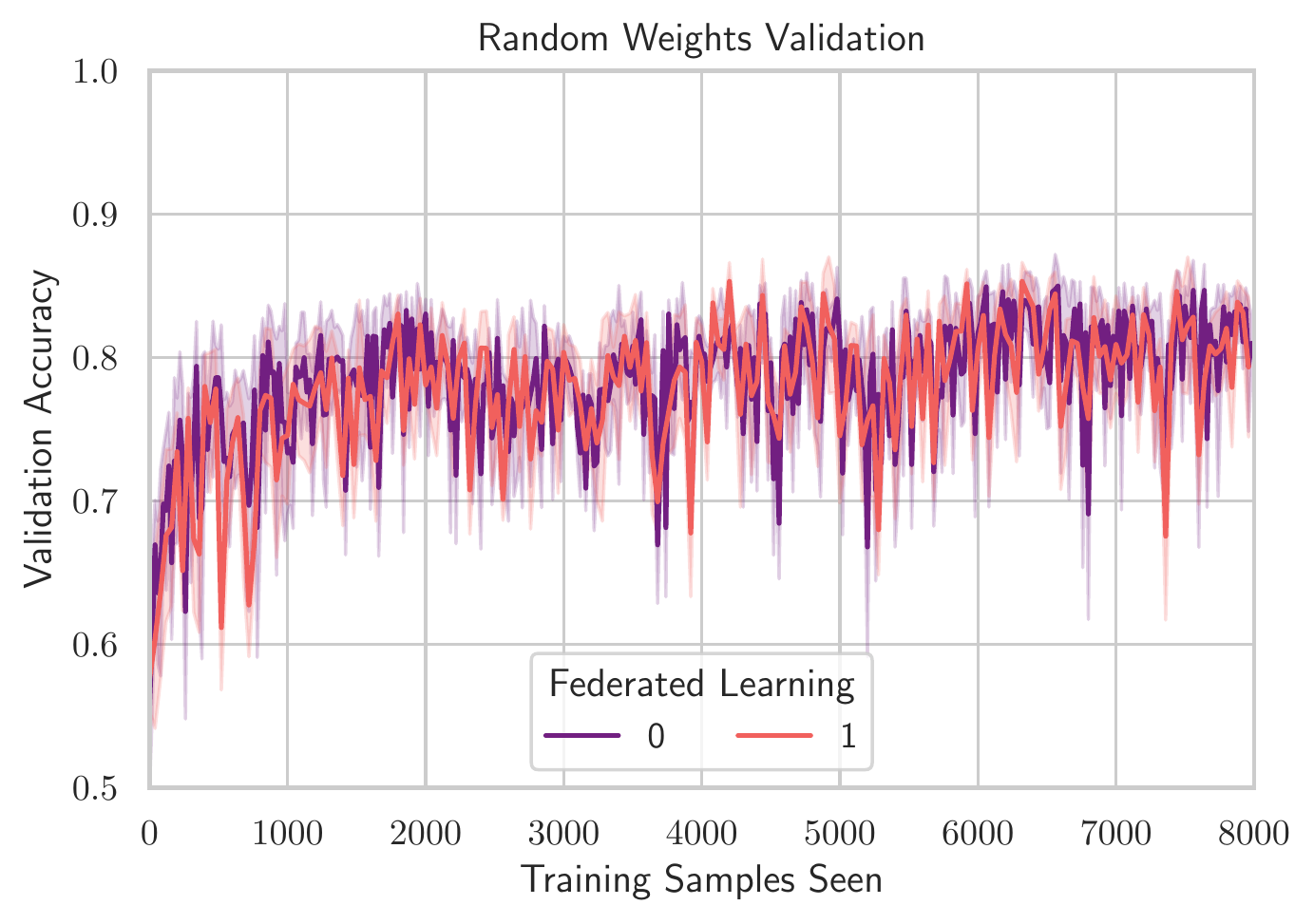}}
\centerline{\includegraphics[width=8.5cm]{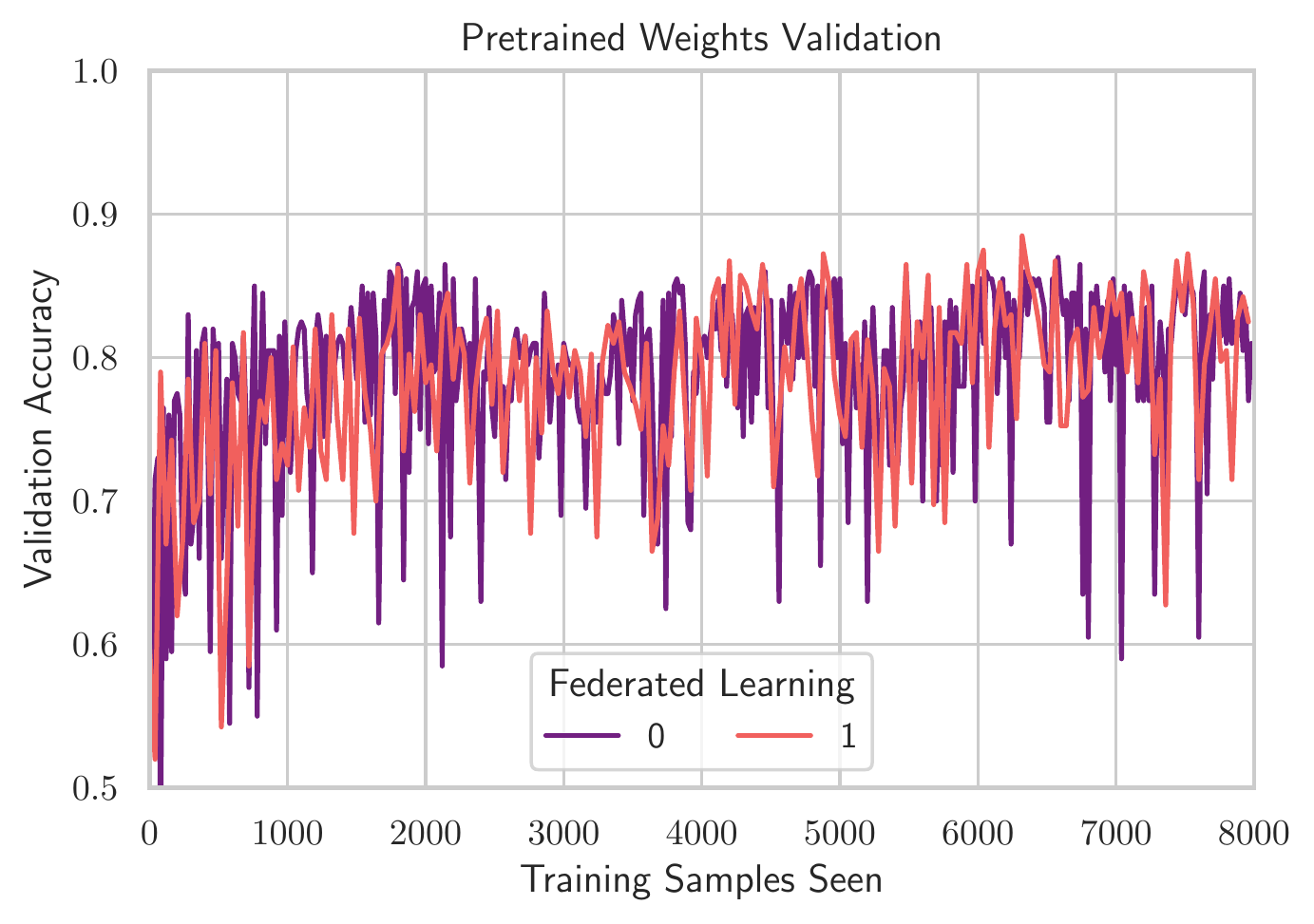}}
\caption{Train and validation accuracy across 10 experiments comparing the training performance of the tf-mobilenet and perf-mobilenet models. The results represent the mean $\pm$ standard deviation across 10 experiments with 5 local episodes and a batch size of 20. The FL implementation has 2 devices. }
\label{fig-perf_tf}
\end{figure}

There are two methods of transfer learning under our setup: using a FC layer trained on a different task and changing the weights, or starting with a randomly-initialized FC layer. Ideally, the differences would be negligible or the randomly-initialized layer would perform better, since then a CNN-based feature extractor could be developed and deployed once and then the size of the dense layers (embedding size x class number) can be varied for various types of applications with different class sizes. As we can see in Figure \ref{fig-perf_tf}, the differences between the randomly initialized model and the model pre-trained on the Dog classification problem have equivalent performance in the Cat classification problem. This is a great result, because it means that whenever we want to switch our feature extractor to a different task to transfer learn, we can just initialize a random FC layer, and the features that were lost in the replacement are not useful for transfer learning.

\subsection{Federated Learning}
Just to note on the following graphs, the measure training examples seen is across all the devices. Therefore, both devices in a federated context and devices not in  federated context are trained and benchmarked against each other. In addition, these results were collected through simulation of the same C++ code that was deployed on the Arudino platform: so the results are identical to performance on-device without having to wait for training and transfer of weighs to occur.  

It's interesting to see that accross all our graphs, models did not benefit from additional data after $\sim 3000$ training examples likely because they were stuck in local optima. This is likely due to the simplicity in the optimizer we used in training our model and presents significant future opportunity.
\subsubsection{Number of Devices}
\begin{figure}[htbp]
\centerline{\includegraphics[width=8.5cm]{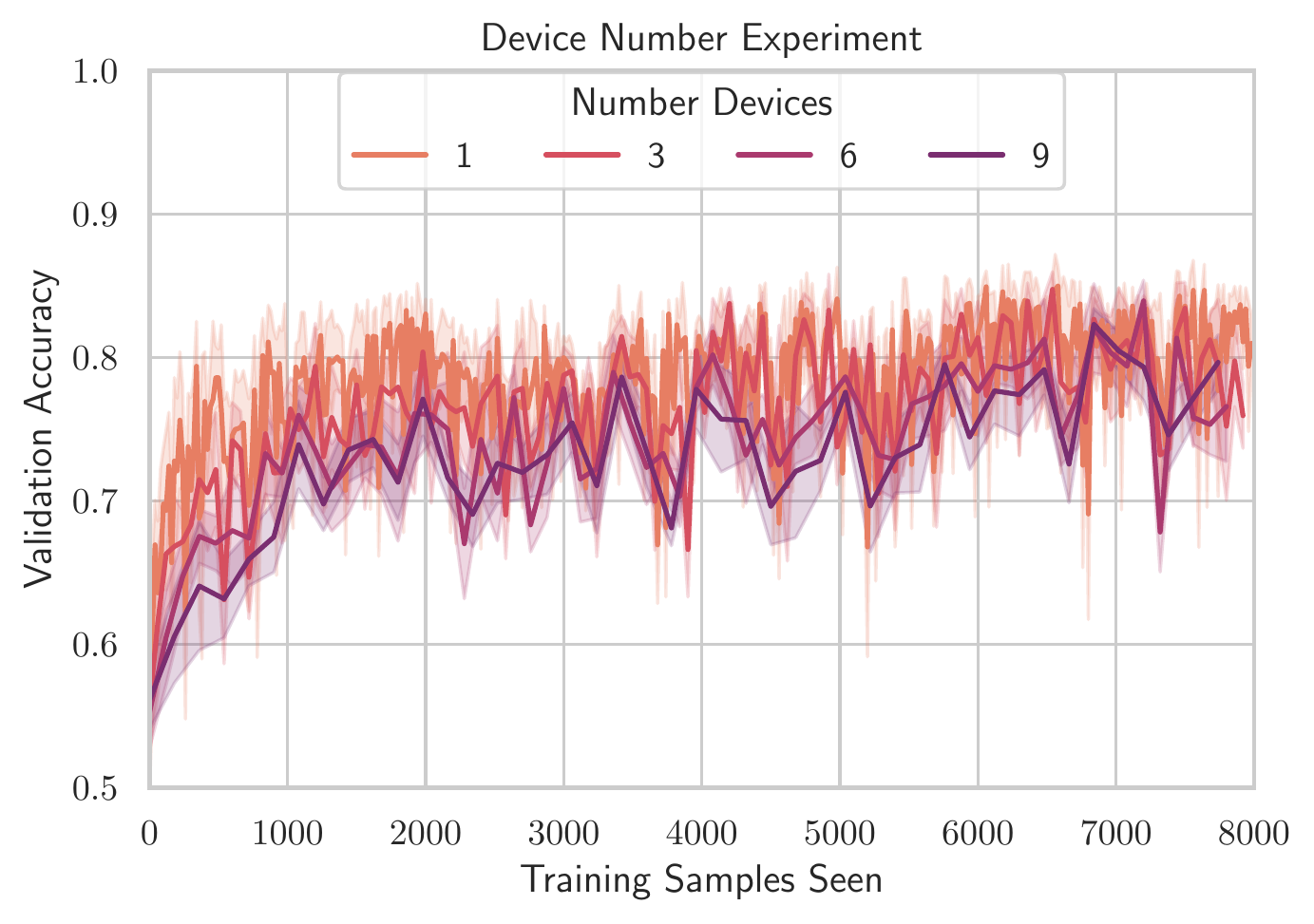}}
\caption{The validation accuracy across epochs of training across different number of devices participating in federated learning. The results represent the mean $\pm$ standard deviation across 10 experiments with 5 local episodes and a batch size of 20.}
\label{fig-dev}
\end{figure}

Figure \ref{fig-dev} shows the effectiveness of federated learning. As we can see, the validation accuracy drops as the number of devices increases, as with more distributed data model updates may cancel each other out or progress may be lost during the averaging step. This is as compared to the baseline of regular (non-FL) transfer learning, shown on the graph as device number of 1.   

\subsubsection{Batch Size}
\begin{figure}[htbp]
\centerline{\includegraphics[width=8.5cm]{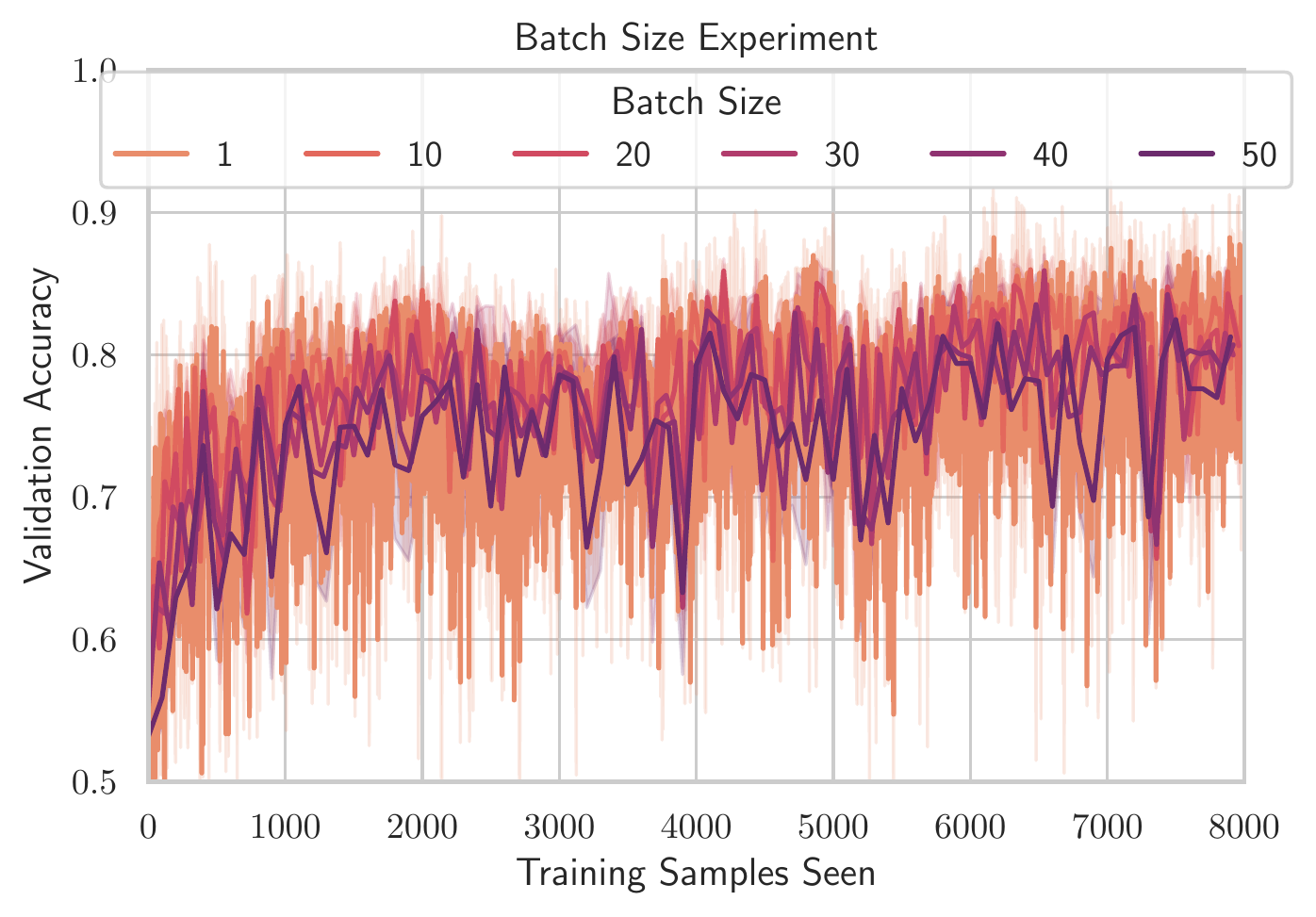}}
\caption{The validation accuracy across epochs of training across different batch sizes. The results represent the mean $\pm$ standard deviation across 10 experiments with 5 local episodes and 2 devices.}
\label{fig-batch}
\end{figure}

As we can see in Figure \ref{fig-batch} that batch size has a significant impact on the variability in epoch-to-epoch performance of FL. This makes sense even in a non-FL scenario, as more data during an update better approximates the true gradients in stochastic gradient descent and prevents over-indexing on a certain piece of data. Ideally in a TinyML application, we would use a batch size of 1 since inference would be instantaneous rather than having to allocate storage and produce results once sufficient data has been collected. 

\subsubsection{Local Episodes}
\begin{figure}[htbp]
\centerline{\includegraphics[width=0.5\textwidth]{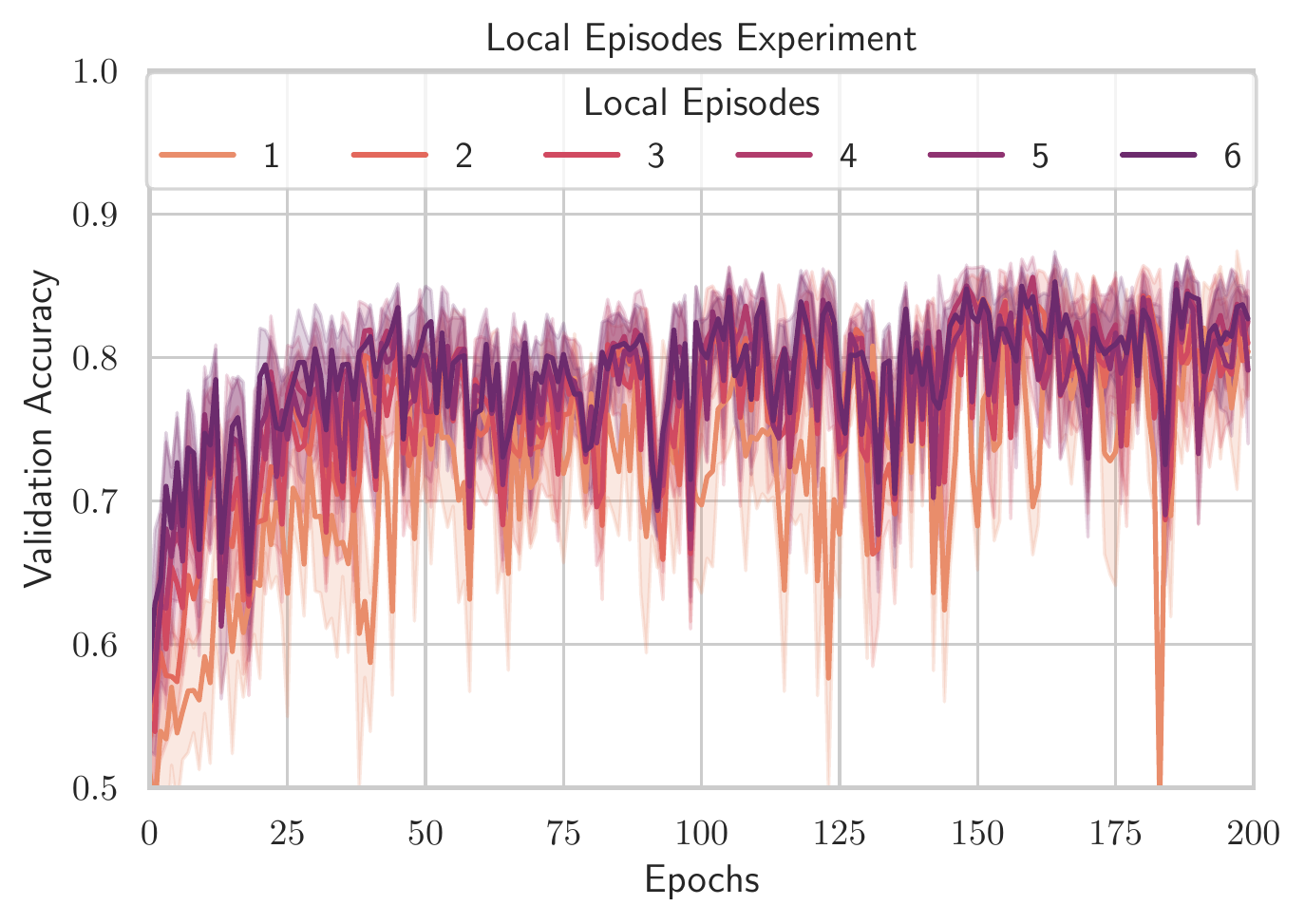}}
\caption{The validation accuracy across epochs of training across different number of local episodes. The results represent the mean $\pm$ standard deviation across 20 experiments with a batch 20 and 2 devices.}
\label{fig-episodes}
\end{figure}

The number of local episodes is defined as the number of times the model was trained on a given batch of data. In Figure \ref{fig-episodes}, it's clear to see that the number of local episodes leads to smaller variability in the epoch-to-epoch validation accuracy. On the other hand, extra local episodes may lead to overfitting to the specific epoch and has greater associated computational costs. For this task, the increase in the number of local episodes from 5 to 6 doesn't present a significant increase in validation accuracy or a decrease in variability. 

\subsection{Performance on-Device}
As stated above, the perf-mobilenet model was used to measure on-device memory and communication cost time. Images were collected through the 5MP arducam, converted to 96x96x3 RGB data, and trained on for 20 local epochs.
\subsubsection{Memory}
Deploying the model utilized 210 KB (80\%) of dynamic memory and 657 KB (66\%) of program storage space. This excludes further memory costs associated with communicating with the global server and encoding captured images.

An important advantage of federated learning is the ability for the device to continuously take in new input data then discard it after it's done training. Since data is not transferred from the edge device to a central server, the device only has to retain its model weights in the federated learning scheme. Therefore, memory footprint does not increase with the number of training samples per device. This is particularly useful in scenarios where external data storage (even SD cards) are not available. Compared to implementations in previous work with KNN and other similar models, we clearly see a vast improvement in memory footprint.
\subsubsection{Time}
By picking a sparse model and writing implementation from scratch without use of external libraries, we were able to optimize our deployment runtime. Image capture and inference took between 8-10 seconds as a buffer from Arducam data was processed as input to the perf-mobilenet model. The actual training process was quite fast as it took only 214ms to go through 20 episodes of local epochs on-device. Then, weights and bias data had to be sent between the arduino and the global server in both directions. These were encoded as bytes, then sent through the serial port in packets of 32 bits. The 514 floats (256x2 weights and 2 bias) often took up on average $\geq$ 6000 bytes, and thus took over 30 seconds (one way) to upload and download.

\section{Discussion}

\subsection{Responsible AI and Privacy}
Our work has shown the possibility of utilizing a federated learning schema to learn tasks in a tinyML domain without the need to share data with a central server. Since data never leaves the device, this opens up avenues for a multitude of applications that have been thus far hesitant to adopt machine learning techniques due to privacy concerns. For instance, in the healthcare context, patient wearables and sensors can continuously assess situations and learn without the risk of privacy infringement for the user. This opens up data in many new fields to be utilized to improve quality of life and user experience without sacrificing privacy.  

Moreover, we show that privacy-centric on-device transfer learning is not only possible but also effective. In our results above, we see that tinyFedTL performed on par with our simulations with regular learning techniques. The federated averaging method effectively captures learnings from edge devices through only sharing of the weights without the need for any attributes of the data.

\subsection{Future opportunities for On-Device Learning}
There are a myriad of future directions this work can go. The first is to implement support for better weight decay and a better optimizer to solve the local optima issue discussed earlier. The second is more algorithmic: in order to further compress the amount of memory needed to run the FC layer training on-device, we could implement feature reduction methods between the embedding from the CNN and the FC layer trained on-device. This would significantly reduce the number of weights stored and trained, therefore also increasing latency and allowing the storage of embeddings to form batches. 

Communication cost is also an area of improvement. Although a 1-minute round-trip communication time per update is not infeasible, this presents problems for edge devices in remote areas with unstable wifi. There, satellites may only have clear connection at short intervals, and packets may be dropped without notice. Thus, possible future work will be to modify our communication protocol for greater resiliency. Other work may take advantage of the sparsity of the model to develop a more efficient message encoding rather than directly translating all weight parameters into bytes. However, there are some costs and decreases in accuracy as the number of devices grow in the federated learning scheme, as shown above. This can be mitigated with capturing more data (our data was limited by the number of images per class in CIFAR-10) and also employing other federated learning techniques like hierarchical training. In those schemas, instead of only having one global server, several intermediary nodes are used to break up edge devices into different groups, meaning there are fewer devices per server node. This would decrease the likelihood of conflicting updates and help sustain accuracy.

\section{Conclusion}
We have shown a successful first-attempt at deploying on-device federated learning. We have demonstrated the efficacy of tinyFedTL with one-shot examples such that storage costs don't increase as the number of training examples increase. We have also shown that pre-trained weights aren't necessary for transfer learning, meaning that edge devices can be re-trained for multiple different classification problems. Finally, our work has also identified several challenges and future avenues of research. The lack of precedent in this task means there is ample room for exploration in terms of designing better optimizers and hypertuning parameters.

\bibliographystyle{unsrt}
\bibliography{references}

\section*{Appendix A: High Level Description of Code}
All our code, including model modifications, C++ NN and FL implementations, Arduino modifications, and more is available at \url{https://github.com/kavyakvk/TinyFederatedLearning}. Below is information on our file structure, graph and result generation code, and the important files for training (marked with a ***).
\begin{verbatim}
dl
source
----arduino_training_final_v3
------->***the .ino file has the implementation 
        of our FL code for the Arduino IDE to 
        compile 
------->***python_final_script.py acts as the 
        "central server" for the arduino
----simulation
------->***NeuralNetwork.cpp has our FC 
        implementation and the FL implementation
------->***simulation.cc is the file with the 
        code necessary to run our simulations.
------->simulation-xxx the executable that can 
        be run with ./ for each of our 
        experiments
-------> the .txt files are the output from 
        terminal when running the experiments
------->fl_simulation_analysis generates the 
        .csv from the .txt
------->graphing.ipynb has the information to 
        graph our figures from the paper 
        from the .csv files
tensorflow (no changes)
third_party (no changes)
\end{verbatim}
\end{document}